\documentclass[12pt]{article} 

\usepackage[utf8]{inputenc} 
\usepackage{amsmath} 
\usepackage{graphicx} 
\usepackage{hyperref} 
\usepackage{geometry} 
\usepackage{multirow}
\usepackage{xcolor}

\title{\textbf{NERCat: Fine-Tuning for Enhanced Named Entity Recognition in Catalan}} 
\author{%
    \small
    \textbf{Guillem Cadevall Ferreres, Marc Serrano Sanz, Marc Bardeli Gámez,} \\
    \small
    \textbf{Pol Gerdt Basullas, Francesc Tarres Ruiz, Raul Quijada Ferrero} \\
    \small
    \href{https://www.ugiat.com/}{Ugiat Technologies} \\
    \small
    \href{https://github.com/ugiat/NERCat}{https://github.com/ugiat/NERCat}
}
\date{\today} 

\begin{document} 

\maketitle 

\begin{abstract}
    Named Entity Recognition (NER) is a critical component of Natural Language Processing (NLP) for extracting structured information from unstructured text. However, for low-resource languages like Catalan, the performance of NER systems often suffers due to the lack of high-quality annotated datasets. This paper introduces NERCat, a fine-tuned version of the GLiNER\cite{gliner} model, designed to improve NER performance specifically for Catalan text. We used a dataset of manually annotated Catalan television transcriptions to train and fine-tune the model, focusing on domains such as politics, sports, and culture. The evaluation results show significant improvements in precision, recall, and F1-score, particularly for underrepresented named entity categories such as Law, Product, and Facility. This study demonstrates the effectiveness of domain-specific fine-tuning in low-resource languages and highlights the potential for enhancing Catalan NLP applications through manual annotation and high-quality datasets.
\end{abstract}


\hrulefill 

\section{Introduction} 
Named Entity Recognition (NER) is a fundamental task in Natural Language Processing (NLP) that involves identifying proper nouns such as persons, organizations, and locations. While widely studied in high-resource languages, NER remains challenging for low-resource languages like Catalan due to limited annotated data. This often leads to errors in distinguishing named entities from common nouns.

This paper presents the fine-tuning of GLiNER, a pre-trained NER model, to improve its performance on Catalan texts. Using a manually annotated dataset derived from Catalan television transcriptions, we enhanced the model’s ability to accurately classify named entities while reducing misclassification errors. The results demonstrate significant improvements in recognition accuracy, contributing to the advancement of Catalan-language NLP applications.

\section{State of the Art} 

NER has advanced from rule-based and statistical methods to modern deep learning and multilingual models. While these approaches show promise, NER for low-resource languages like Catalan faces challenges due to limited annotated data and the dominance of high-resource languages in multilingual models. This section reviews these methods, their strengths, and their limitations for Catalan:

\begin{itemize}
    \item \textbf{Rule-Based and Statistical Methods:} Early NER systems relied on handcrafted rules and statistical models (e.g., Hidden Markov Models and Conditional Random Fields). While effective for structured data, these approaches struggle with complex sentence structures and linguistic variability.
    \item \textbf{Supervised Deep Learning Models:} Modern neural architectures, such as BiLSTMs and Transformer-based models (e.g., BERT, XLM-R), have significantly improved NER performance. However, their effectiveness is dependent on the availability of high-quality annotated datasets, which are scarce for Catalan.
    \item \textbf{Multilingual Pre-Trained Models:} Recent advancements in multilingual NLP have led to models like mBERT, XLM-R, and GLiNER, which leverage cross-lingual knowledge transfer. While these models offer baseline support for Catalan, they often suffer from inaccuracies due to the dominance of high-resource languages in their training data.
\end{itemize}

Despite these advancements, existing NER models frequently misclassify entities, particularly in mixed-language contexts where Catalan coexists with Spanish. Additionally, domain-specific named entities (e.g., Catalan media, politics, and culture) are often underrepresented, limiting model generalization. To overcome these challenges, this study adopts a customized approach aimed at improving entity recognition, as detailed in the following section.

\section{Methodology} 
Our approach to improving NER in Catalan involved fine-tuning GLiNER Knowledgator\cite{knowledgator}, a pre-trained Transformer-based model, using a high-quality annotated dataset. This section describes the model selection, dataset preparation, annotation process, and fine-tuning procedure.

\subsection{Model Selection} 
For this study, we selected GLiNER Knowledgator, a Transformer-based NER model that has shown strong performance in languages other than English. However, its baseline performance in Catalan was poor, primarily due to a lack of dedicated training data. A major issue with GLiNER Knowledgator is its tendency to misclassify common nouns as named entities, leading to a high false positive rate. This issue is particularly pronounced in Catalan, where proper names and common nouns often share morphological similarities, making entity differentiation challenging.

To address these limitations, we fine-tuned GLiNER Knowledgator on a manually annotated Catalan dataset, focusing on improving its ability to correctly identify named entities while reducing misclassification errors.

Additionally, to build a rich and diverse dataset, we employed Whisper Fast\cite{whisper}, an automatic speech-to-text model, to extract transcriptions from Catalan-language television programs. This provided real-world Catalan text samples that better reflect actual language use across multiple domains.

The specific pre-trained version was: 'knowledgator/gliner-bi-large-v1.0'.

\subsection{Dataset Preparation} 
The dataset used for fine-tuning was constructed from Catalan-language television transcriptions to ensure broad coverage of real-world text. The preparation process included multiple steps to refine data quality and enhance model training. Key characteristics of the dataset are:

\begin{itemize}
    \item \textbf{Source:} Transcriptions extracted from Catalan television programs, ensuring diverse linguistic coverage.
    \item \textbf{Speech-to-Text Model:} Whisper Fast was employed for automatic transcription, providing a solid baseline for further annotation.
    \item \textbf{Language Composition:} The dataset primarily consists of Catalan-language texts, with occasional Spanish code-switching segments.
    \item \textbf{Size:} A total of 9,242 manually annotated sentences, covering various domains such as politics, sports, and culture, along with 13,732 NERs.".
    \item \textbf{Annotation Process:} Human experts carefully labeled named entities to enhance accuracy and reduce ambiguities.
    \item \textbf{Purpose:} This dataset was designed to support Catalan NLP applications, such as the Projecte AINA\cite{projecte-aina} initiative, by enhancing NER performance in underrepresented linguistic settings.
\end{itemize}

The dataset annotation follows a structured set of entity labels to ensure consistency and accuracy:
\begin{itemize}
    \item \textbf{Person:} Names of individuals, including fictional and historical figures.
    \item \textbf{Facility:} Physical infrastructures such as airports, bridges, and stadiums.
    \item \textbf{Organization:} Entities such as companies, institutions, political parties, and media outlets.
    \item \textbf{Location:} Geographical places, including cities, regions, and landmarks. This also includes names of countries, mountain lakes, or planets.
    \item \textbf{Product:} Commercial products, such as branded phones or soft drinks. This category also includes cultural products, such as the titles of films, books, and songs.
    \item \textbf{Event:} Named occurrences like festivals, elections, or historical battles.
    \item \textbf{Date:} Temporal expressions indicating specific days, months, years, or time periods.
    \item \textbf{Law:} Names of laws, legal frameworks, and judicial matters, including constitutions, court rulings, and high-profile judicial cases.
\end{itemize}

\subsection{Manual Annotation Process} 
To ensure high-quality entity classification, we manually annotated the dataset using the \textit{NER Text Annotator} tool\cite{ner-annotator}. This open-source tool facilitated an efficient and accurate annotation process, allowing experts to systematically label named entities while minimizing misclassifications.

The annotation workflow consisted of three key steps:

\begin{itemize}
    \item \textbf{Sentence Segmentation:} The transcriptions were split into individual sentences for easier processing.
    \item \textbf{Entity Labeling:} Using the NER Annotator interface, human annotators manually assigned entity labels to relevant words.
    \item \textbf{Dataset Compilation:} The annotated dataset was structured into a format compatible with GLiNER fine-tuning.
\end{itemize}

\subsection{Alternative Approaches Explored} 
Before selecting manual fine-tuning as the primary approach, we explored several alternative methodologies to enhance NER in Catalan. However, each of these approaches presented limitations that ultimately led us to prioritize human annotation for optimal accuracy and reliability.

One such approach involved utilizing Large Language Models (LLMs), where we provided various prompts to classify files from Wikipedia in an attempt to automate the entity recognition process. However, this approach yielded suboptimal results. Specifically, the LLMs had significant difficulty distinguishing between proper nouns and common nouns, leading to erroneous classifications. As a result, the performance was not sufficient for our needs, and the approach was deemed inadequate for achieving the desired accuracy. These limitations ultimately led us to prioritize human annotation, which provided the necessary accuracy and reliability for NER tasks in Catalan.

\subsection{Fine-Tuning Process} 
The fine-tuning process for GLiNER in Catalan NER followed a structured approach, including dataset preparation, model training, and optimization. To begin, the annotated dataset was shuffled to eliminate order bias and then split into training (90\%) and testing (10\%) subsets, ensuring a balanced evaluation. The fine-tuning was performed on the pre-trained \texttt{knowledgator/gliner-bi-large-v1.0} model.

For training, the model used a batch size of 8, applying focal loss ($\alpha = 0.75, \gamma = 2$) to address class imbalances. Learning rates were set at $5 \times 10^{-6}$ for entity layers and $1 \times 10^{-5}$ for other model parameters, with a linear scheduler incorporating a warmup ratio of 0.1. The model was evaluated every 100 steps to track performance improvements.

The dataset contained 13,732 named entity instances across multiple categories:

\begin{itemize}
    \item \textbf{Person}: 5459 (37.90\%)
    \item \textbf{Facility}: 496 (3.44\%)
    \item \textbf{Organization}: 3123 (21.68\%)
    \item \textbf{Location}: 2840 (19.72\%)
    \item \textbf{Product}: 580 (4.03\%)
    \item \textbf{Event}: 478 (3.32\%)
    \item \textbf{Date}: 636 (4.42\%)
    \item \textbf{Law}: 120 (0.83\%)
    \item \textbf{No label}: 673 (4.67\%)
\end{itemize}

The high presence of "Person" and "Organization" entities reflects the dataset’s origin in television transcriptions, where political and cultural discussions dominate. The fine-tuning process aimed to balance recognition accuracy across all categories while minimizing misclassification errors, particularly in underrepresented types such as "Law" and "Facility."

\section{Results and Analysis} 

The evaluation of our fine-tuned NERCat model was conducted using a manually classified dataset of 100 sentences, specifically curated as an evaluation dataset. The primary goal was to compare its performance against the baseline GLiNER model and analyze improvements across various named entity categories.

Table \ref{tab:comparison_results} presents the comparative performance metrics, including precision, recall, and F1-score, for both NERCat and GLiNER across eight entity types: Person, Facility, Organization, Location, Product, Event, Date, and Law. The results clearly indicate that fine-tuning significantly enhances entity recognition, particularly in underrepresented categories.
\begin{table}[h]
    \centering
    \resizebox{\textwidth}{!}{
    \begin{tabular}{|c|c|c|c|c|c|c|c|c|c|}
        \hline
        \multirow{2}{*}{\textbf{Entity Type}} & \multicolumn{3}{c|}{\textbf{NERCat}} & \multicolumn{3}{c|}{\textbf{GLiNER}} & \multicolumn{3}{c|}{\textbf{Improvement ($\Delta$)}} \\
        \cline{2-10}
        & \textbf{Prec.} & \textbf{Rec.} & \textbf{F1} & \textbf{Prec.} & \textbf{Rec.} & \textbf{F1} & \textbf{$\Delta$ Prec.} & \textbf{$\Delta$ Rec.} & \textbf{$\Delta$ F1} \\
        \hline
        Person & 1.00 & 1.00 & 1.00 & 0.92 & 0.80 & 0.86 & \textbf{+0.08} & \textbf{+0.20} & \textbf{+0.14} \\
        Facility & 0.89 & 1.00 & 0.94 & 0.67 & 0.25 & 0.36 & \textbf{+0.22} & \textbf{+0.75} & \textbf{+0.58} \\
        Organization & 1.00 & 1.00 & 1.00 & 0.72 & 0.62 & 0.67 & \textbf{+0.28} & \textbf{+0.38} & \textbf{+0.33} \\
        Location & 1.00 & 0.97 & 0.99 & 0.83 & 0.54 & 0.66 & \textbf{+0.17} & \textbf{+0.43} & \textbf{+0.33} \\
        Product & 0.96 & 1.00 & 0.98 & 0.63 & 0.21 & 0.31 & \textbf{+0.34} & \textbf{+0.79} & \textbf{+0.67} \\
        Event & 0.88 & 0.88 & 0.88 & 0.60 & 0.38 & 0.46 & \textbf{+0.28} & \textbf{+0.50} & \textbf{+0.41} \\
        Date & 0.88 & 1.00 & 0.93 & 1.00 & 0.07 & 0.13 & \textbf{-0.13} & \textbf{+0.93} & \textbf{+0.80} \\
        Law & 0.67 & 1.00 & 0.80 & 0.00 & 0.00 & 0.00 & \textbf{+0.67} & \textbf{+1.00} & \textbf{+0.80} \\
        \hline
    \end{tabular}
    }
    \caption{Comparison of performance metrics (Precision, Recall, and F1-score) between NERCat and GLiNER across different entity categories. The last three columns show the improvement ($\Delta$) in each metric.}
    \label{tab:comparison_results}
\end{table}

\subsection{Performance Comparison}

The NERCat model outperforms GLiNER in nearly all categories, achieving almost perfect precision and recall scores in Person, Organization, and Location entities. However, the greatest improvements are seen in categories with previously poor recognition, such as Facility, Product, and Law, where F1-scores increased by 57.75\%, 66.71\%, and 80.00\% respectively. These improvements highlight the effectiveness of the manually annotated dataset in refining model performance.

\subsection{Error Analysis}

While NERCat demonstrates significant enhancements, some challenges remain. For instance, the Date category shows a 12.5\% drop in precision; however, this is due to GLiNER's recall being only 7.14\%, meaning it detects almost no date-related entities, leading to an artificially high precision of 100\%. In contrast, NERCat achieves an 80\% recall, demonstrating a substantial improvement in entity detection despite a slight trade-off in precision.

\subsection{Impact on Catalan NLP}

The results from this evaluation underscore the importance of fine-tuning pre-trained models with domain-specific and manually annotated datasets. By leveraging real-world Catalan television transcriptions, NERCat has substantially improved entity recognition accuracy, paving the way for better NLP applications in Catalan media, governance, and cultural domains.

Overall, our findings demonstrate that targeted fine-tuning, combined with high-quality manual annotations, is crucial for advancing NLP tools in low-resource languages like Catalan. Future work will focus on expanding the dataset and exploring additional linguistic variations to further refine entity recognition capabilities.

\end{document}